# Analysis of child development facts and myths using text mining techniques and classification models

Mehedi Tajrian [a,*], Azizur Rahman [a], Muhammad Ashad Kabir [a,*], and Md Rafiqul Islam [a]

[a] *School of Computing, Mathematics and Engineering, Charles Sturt University, NSW, Australia*
[*]Corresponding authors.

## ABSTRACT

The rapid dissemination of misinformation on the internet complicates the decision-making process for individuals seeking reliable information, particularly parents researching child development topics. This misinformation can lead to adverse consequences, such as inappropriate treatment of children based on myths. While previous research has utilized text-mining techniques to predict child abuse cases, there has been a gap in the analysis of child development myths and facts. This study addresses this gap by applying text mining techniques and classification models to distinguish between myths and facts about child development, leveraging newly gathered data from publicly available websites. The research methodology involved several stages. First, text mining techniques were employed to pre-process the data, ensuring enhanced accuracy. Subsequently, the structured data was analysed using six robust Machine Learning (ML) classifiers and one Deep Learning (DL) model, with two feature extraction techniques applied to assess their performance across three different training-testing splits. To ensure the reliability of the results, cross-validation was performed using both k-fold and leave-one-out methods. Among the classification models tested, Logistic Regression (LR) demonstrated the highest accuracy, achieving a 90% accuracy with the Bag-of-Words (BoW) feature extraction technique. LR stands out for its exceptional speed and efficiency, maintaining low testing time per statement (0.97 microseconds). These findings suggest that LR, when combined with BoW, is effective in accurately classifying child development information, thus providing a valuable tool for combating misinformation and assisting parents in making informed decisions.

## 1. Introduction

In the internet age, misinformation spreads rapidly, leading to anxiety, depression, and emotional exhaustion among individuals [1, 2]. Awareness of misinformation is crucial to prevent being misled. Myths, a form of misinformation, are often false statements that go unchallenged and are perpetuated through social and familial influences [3]. Although some myths may hold true for a few individuals, they are generally misunderstood by the broader public. Research on child development is vital [4]. Proper treatment of children is essential for their optimal growth and development. Parents' decisions about their children's nutrition and overall development are often influenced by the information they find online [5]. Unfortunately, not all this information is accurate, and misinformation can lead to adverse effects on children's well-being [6].

A critical gap exists regarding the comprehensive analysis of myths and facts surrounding child development. Data on child development myths and facts have not yet been analysed using modern data science techniques. A few studies used text mining to predict child abuse cases [7, 8]. This study aims to apply text mining techniques and classification models to analyse and differentiate myths and facts about child development. By leveraging newly gathered data from publicly available websites, the study seeks to provide valuable insights and improve the accuracy of information parents use in decision-making.

The research addresses the following questions: (i) how effective are text mining techniques in pre-processing and structuring data on child development myths and facts? (ii) which classification models perform best in terms of accuracy when applied to this data? (iii) how do different feature extraction techniques impact the performance of these models? (iv) what validation methods are most reliable for verifying the accuracy of these models? and (v) how cost-effective are these models?

The key contributions of this study are fourfold: (i) providing a comprehensive analysis of the new data, that is made usable through several text mining techniques including categorization, preprocessing and cleaning, (ii) analysing the data by feature extraction techniques and classification strategies, and comparing with deep learning method for different training-testing sizes; (iii) verifying the analysis results through several validation methods; and (iv) evaluating the cost-effectiveness of the models. The findings from this study would enrich the literature of data science which will help the researchers for their future analysis. It will also add a new domain of research findings about child development facts and myths, from which the parental community would be able to take better decisions for their children.

This study applies text mining techniques to pre-process the data for enhancing the performance and accuracy. The structured data is then analysed using six classification and one deep learning models with two feature extraction techniques for three training-testing splits. Cross-validation is performed to ensure the reliability of the results. The findings indicate that LR achieves the highest accuracy at 90% using BoW. LR excels in speed and efficiency by maintaining low testing time per statement (0.97 µs) for both feature extraction methods. The results demonstrate that LR is effective in classifying child development information, providing a valuable tool for debunking myths and aiding parents in making informed decisions. The findings enrich the data science literature and open a new domain of research on child development myths and facts, ultimately helping the parental community make better decisions for their children.

The rest of the paper is structured as follows. Section 2 covers the background and motivation of the study, along with a review of related research. Section 3 details the methodology for detecting facts and myths. Section 4 describes the experimental evaluation, while Section 5 presents the results of our evaluation. The key findings of the study are discussed in Section 6. Finally, Section 7 concludes the paper.

## 2. Background and related work

Misinformation must be defined properly before being investigated to gain perfect accuracy according to the analysis objective. Examining recent research this study explains the background of misinformation and myths, as well as the motivation behind this research.

### *2.1. Background*

Misinformation aims to mislead people or gain their attention [8-10]. The term misinformation refers to "false, mistaken, or misleading information" [11] or "any information deemed false due to a lack of scientific evidence" [12]. [13] addresses misinformation, explains its impact, provides insight into India's context, and discusses vaccine misinformation. Misinformation has been associated with terms like disinformation, misleading information, fake news, maliciously false news, false news, satire news, conspiracy theories, and rumour. There have also been studies exploring social spam, fake reviews, and clickbait as forms of misinformation on the internet. As myths, misinformation is passed down through the generations.

There is a strong correlation between disinformation and misinformation [14, 15]. In contrast to misinformation, disinformation refers to misleading information [16]. For example, it perpetuates misconceptions, such as that alcohol kills Coronaviruses [17], harming the health of those who consume it. Usually, misleading information is provided by eyewitnesses after an event [18]. It has been established that misleading information concerning COVID-19 [19] can modify economies, erode individual trust in governments, or greatly benefit the country.

Rumours are unconfirmed information from official sources [20]. Most rumour spreads on social media platforms [21, 22]. Rumours are circulating stories of uncertain veracity that are credible, but hard to verify and cause skepticism and/or anxiety [23, 24]. Rumours can be classified according to their characteristics, scopes, and types. Assesses rumours about expected reactions by using a psychological approach [25]. Rumours can be classified as long-standing or short-lived, depending on whether they were verified as true or false over the long term. Urban legends and conspiracy theories are some types of long-standing rumours. There are also breaking news rumours, which may be unintentional misinformation but are also misleading. The spread of rumours can be prevented if they are identified quickly, especially if they have malicious intent.

Conspiracy theories are believed to be created by secret or powerful groups [26-28]. They operate over the Internet to harm people [28, 29]. During times of crisis, including financial emergencies, diseases, natural disasters, terrorist attacks, and wars, people believe in them [30-32]. Several conspiracy theories have emerged in response to COVID-19's turmoil; for instance, “5G cellular networks are at the root of the virus” and “Bill Gates is using the virus as a cover for his desire to establish a worldwide surveillance state by enforcing a global vaccination program” [33].

The definition of fake news remains unclear [34-37]. Researchers define fake news as "news articles that are intentionally and verifiably false" [38, 40]. In fake news, readers are intentionally misled or misinformed [21, 41-44]. Various fake news exists, including large-scale hoaxes, serious fabrications, and clever fabrications. In general, fake news is propagated via social media with malicious intent, such as articles with malicious intent. False information presented as good news is the hallmark of large-scale hoaxes. Usually, a hoax against a public figure or an idea is more elaborately arranged than simple news reports. Readers know the author's humorous intentions when clever fakes are created. Some humorous articles are masquerading as news, including those published by The Onion and lercio.it [21, 41, 45]. The author of [45] presents a review of existing methods and techniques for detecting fake news. Several state-of-the-art technologies, models, datasets, and experimental setups are discussed in [46]. Using clickbait in article titles or social media posts is a common practice to encourage visitors to click links to article pages [47]. Internet fake news is largely spread through clickbait headlines [12]. Media review aggregators and e-commerce websites are the most common places to find fake reviews [21]. Spam messages often contain malicious links, apps, fake accounts, fake news, reviews, rumours, etc. [48, 49].

Myths are stories passed down through generations, often with elements of fantasy and some elements of truth. The search for knowledge or understanding and the struggle between good and evil are common motifs in myths. Child development myths and misconceptions must be understood to make informed decisions regarding how to raise and nurture children. Parents and caregivers can better understand how to create an environment for healthy development by addressing several common myths about child development. It is a myth that a child's environment does not influence his or her development. Parents and caregivers must provide children with a stimulating, positive environment that encourages them to explore and learn.

We must ensure the optimal development of our children to ensure the future of our world. Parental decisions (for example, about their children's diet and nutrition) are influenced by free and readily available information online, including myths and facts about child development. Modern parents often search online for information about things they do not know or want to learn in today's data-centric world. Information available on the internet is not always accurate, and some information might be considered misinformation. Children can be adversely affected by myths if they are treated according to them, and parents may become confused when they know the myths.

### *2.2. Related work*

Child development is a topic of great interest and importance in understanding the growth and well-being of children. However, amidst the wealth of information available, misconceptions and myths often cloud our understanding of what truly influences a child's development. Despite the myths and facts associated with child development, a direct study has not been conducted.

The research article [7] has been found as closely related to one of our research objectives: text mining. Text mining analysis was used to identify and predict child abuse cases in a public health institution in the Netherlands. They collected medical data as unstructured free text, investigated the data, and extracted valuable patterns. Using machine learning models, they achieved a high score in classifying cases of possible abuse and implemented a decision support Application Programming Interfaces (API) at a municipality in the Netherlands.

Considering many challenges that most societies face, the study [50] aims to identify predators during the very early stages of online communication. By using Machine Learning techniques, they investigated how to detect child grooming from online chat records. This study was able to make a multi-label classification based on toxicity types on a Wikipedia dataset with more than 97 percent accuracy. PAN12 dataset was used for training and testing our model. More than 92 percent of the suspicious conversation messages from the chat records can be identified, and sexual predators can be identified.

The article [51] provides an inside look at predictive tools by showing how children in a population are most likely to be maltreated, and how they can be developed for social work, given the data recorded by machine learning about service users and service activity. As part of the ongoing reforms to child protection services, the development of predictive risk modelling (PRM) in New Zealand is considered as an example.

The study [52] created a linear prediction model that had 45.2% sensitivity and 82.4% specificity using six predictors. This model could be useful for assessing the risk of further maltreatment and developing recurrence prevention strategies for child welfare agencies.

The purpose of this study [53] is to assess the evolutionary trends of concerned topics as well as the differences in sentiment expression in topics during elementary school children's development by mining 4556 free-text writing data based on ecosystem theory and topic modelling in text mining [54].

Text mining and machine learning techniques are used in the study [55] to detect Korean fake news. A text mining technique (Topic Modelling, TF-IDF, etc.) is used to transform the news contents to quantified values. n the subsequent step, classifiers are trained using these quantified values. Using Seoul National University's FactCheck, they collected 200 Korean news articles and linked to source documents to test the effectiveness of the proposed method.

Designed to target misinformation about autism interventions in a real-world setting, the study [56] tested optimized debunking strategies. The researchers randomized participants to a “optimized-debunking” or a “treatment-as-usual” condition as part of professional development training and compared the support for non-empirically supported treatments before, after, and 6 weeks after the training.

With a focus on fake reviews and fake news, a new n-gram model was introduced to detect automatically fake contents [57]. There were two different methods for extracting features and six methods for classifying using machine learning. Comparing the state-of-the-art methods with existing public datasets and a new fake news dataset indicates greatly improved performance.

By using ground-truth data, the study [19] built a machine-learning detection system that can detect misleading information. A voting ensemble machine learning classifier is constructed by using ten machine learning algorithms and seven feature extraction techniques. Several performance metrics were evaluated after the data were cross validated five times. According to the evaluation results, the collected ground-truth data are of high quality and validity and are effective for the creation of models to detect misleading information.

Based on content-only features, they [58] reviewed several state-of-the-art Machine Learning (ML) and neural network models. A Term Frequency-Inverse Document Frequency (TF-IDF) feature was used for ML training to enhance classification performance, whereas a word embedding feature was used for neural network training to enhance classification performance. All traditional models are more accurate than 85% when ML and Natural Language Processing (NLP) methods are used. Over 90% accuracy is achieved by all neural network models. As a result of the experiments, they found that neural networks outperformed traditional machine learning models by approximately 6% on average and reached up to 90% accuracy in all cases.

An analysis of words appearing on the first page of the most popular English newspaper in Bangladesh, The Daily Star, in 2018 and 2019 is presented in the paper [59]. Using word patterns, they attempted to explain the possible political and social context of that era. There are three commonly employed and current text mining techniques used in the study: word clouds, sentiment analysis, and cluster analysis.

According to [60], contextual information can improve the detection of hate speech. They examined a subdomain of Twitter data containing replies to digital newspaper posts, which provides a natural environment for the detection of contextualised hate speech. Their interdisciplinary team built and annotated a new corpus in Spanish (Rioplatense variant) that focused on hate speech associated with the COVID-19 pandemic. They found that context improves hate speech detection for binary and multi-label prediction by 4.2 and 5.5 points, respectively, based on their classification experiments using transformer-based machine learning techniques.

A key factor in fake news' virality is surprised, which draws the reader's immediate attention and elicits strong emotional responses. Using this idea, in [61], they proposed textual information detection and sentiment analysis as two tasks related to automatic misinformation detection. Textual entailment is repurposed for novelty detection, and models trained on large-scale datasets of entailment and emotion are used to identify fake information. On four large-scale misinformation datasets, they achieved optimal performance.

A probabilistic approach is used by [62] to assess the accuracy of information. Using NLP-based methods and statistical analysis, the system identifies some features after analysing fake news in the literature. The specified features highlighted the information's syntactic, semantic, and social characteristics. From these features, previously developed on a piece of fake dataset news, a Bayesian Network was constructed to estimate the likelihood of accuracy. It has been tested in some real cases with satisfactory results.

The paper [63] analyses the studies on detecting fake news and investigates the conventional machine-learning approaches that can be used to select the most accurate ones. Using tools such as Python scikit-learn, Natural Language Processing for textual analysis, and Python scikit-learn, a supervised machine learning algorithm can be created to create a model of a product that can classify fake news as true or false. As a result of this process, they proposed to use Python scikit-learn library to implement tokenization and feature extraction; since this library includes efficient tools, such as Count Vectorizer and TF-IDF Vectorizer. Afterwards, they applied feature selection methods to test and select the most appropriate features based on the confusion matrix results.

Various text-mining techniques have been used to uncover common misconceptions. To shed light on these misconceptions and systematically analyse and identify prevailing myths and facts in the field of child development, text-mining techniques have been tuned. One of the primary objectives of text-mining research in child development is to identify prevalent myths that may mislead parents, educators, and caregivers. Primary sources were used in this study to collect data. The primary data were analysed using text mining techniques such as removing numbers, punctuations, stop words, extra spaces, case transferring, tokenization, stemming, lemmatization, word cloud, sentiment analysis. Two feature extraction techniques have been applied on six most robust and suitable classification techniques and a deep learning model for the new data. The accuracy and cost-effectiveness of the machine and deep learning techniques have been checked with some evaluation matrices, a performance indicator and cross validation techniques.

## 3. Methodology

This section aims to discuss the data collection mechanism and methodology applied in this study. The fact and myth data has been collected separately from the websites. Several data cleaning methods and text mining techniques are applied to preprocess the data. Word cloud is used to visualise various insights of the data. Two feature extraction techniques, BoW and TF-IDF, are applied on the data to enrich the accuracy. The data is analysed using various Artificial Intelligent (AI) techniques, including LR, Random Forest (RF), Support Vector Machines (SVM), Naive Bayes (NB), Decision Tree (DT), K-Nearest Neighbours (KNN), and Convolutional Neural Networks (CNN) to find the most suitable one for the new data. Each of these six ML and one DL strategies has been investigated with different training-testing splits. The accuracy was again confirmed using cross-validation techniques. To evaluate the cost-effectiveness of these models a performance indicator is used. The steps in our proposed methodology have been described in Fig. 1.

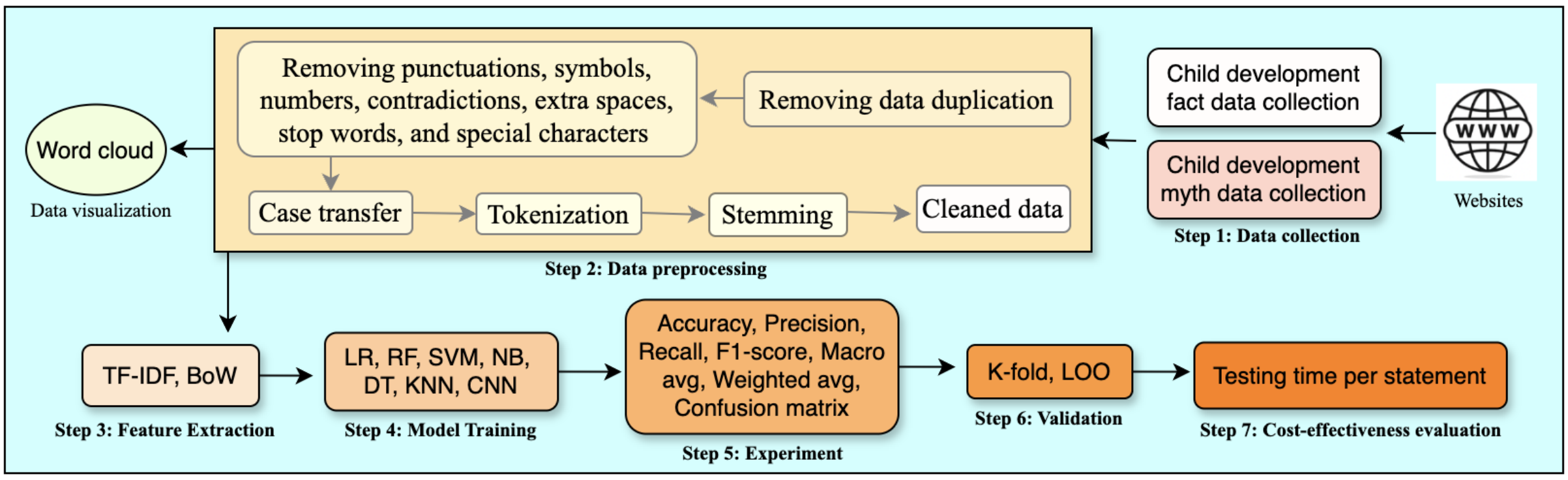

**Figure 1.** Methodological steps.

### *3.1. Data collection and labelling*

The first step in collecting primary data involves searching the internet for facts and myths about children's development. A total of 105 websites were analysed, 65 of which dealt with myths and 40 with facts. In the analysis, there were 436 myths and 953 facts. Further analysis is conducted using these two labels: facts and myths. The following examples are taken from the data presented in Table 1. In the data, there are two columns - labels and statements.

**Table 1.** Dataset with labels and statements fields.

| Label | Statements |
|---|---|
| Fact | Play is like children's work. |
| Fact | Children communicate to express their needs. |
| Fact | Young babies need consistent responsive care. |
| Myth | Lifting weights stunts growth. |
| Myth | It's too late for my kid. |
| Myth | Little Brain, Little Activity. |

### *3.2. Data preprocessing*

Data preprocessing is an essential step in preparing raw data for analysis and modelling. A format is created by transforming raw data that is easier to work with and more suitable for analysis. Data transformation is another key step in data preprocessing. This is accomplished by converting the data into a form that is better suited for analysis. This labelled data is preprocessed by removing punctuation, numbers, stop words, contradictions, symbols, extra spaces, and case transfer. Many text mining techniques are used including tokenization and stemming to find the most meaningful words for further analysis as described in step 2 of Fig. 1.

Tokenization is a technique, implicating which text is broken down into separate words or tokens. Doing so makes it easier to analyse the content and identify unusual or misleading patterns. Stemming and lemmatization involve reducing words to their root form, allowing easier comparison and analysis. By identifying the root form of words used in a piece of information, it becomes possible to identify any attempts to manipulate or mislead using similar or related words.

### *3.3. Word cloud*

As a visual representation of a collection of words, a word cloud represents the frequency or importance of each word in the text. It is an efficient method for analysing and visualizing text data quickly. The frequency of each word is calculated after preprocessing. As a word appears more often, its size will increase in the word cloud. Fig. 2 shows a word cloud that displays the most frequent words in a larger font size, while less frequently occurring words appear in a smaller font size.

### *3.4. Feature extractions*

Feature extraction is the process of selecting and extracting meaningful and relevant features from a given dataset. Machine learning algorithms can be trained and evaluated using these features. The dimensionality of the dataset can be reduced by extracting relevant features that make the classifiers more manageable and efficient. Additionally, feature extraction can help identify patterns that may be too complex or noisy for machine learning algorithms to identify directly. There are several types of feature extraction methods that can be used for different types of data. We have used TF-IDF and BoW techniques as they are best suits for our data analysis, as depicted in step 4 of Fig. 1.

*1) Term Frequency-Inverse Document Frequency (TF-IDF)*: It is a feature extraction technique commonly used for text data. Term frequency in the document and inverse document frequency across all documents are considered. In this measure, unique and important words in a particular document are identified while frequently occurring words are ignored.

*2) Bag of Word (BoW)*: It is a dimensionality reduction method that converts text data into a dense vector representation. Each word in the document is counted according to its frequency and represents each document as a weighted sum of its word frequencies. This representation can capture the document's essence without considering the words' order.

### *3.5. Classifications*

Classification is a type of data analysis technique used to group or categorize items into predefined categories based on their characteristics or features. Based on training data, it identifies patterns and makes predictions, a crucial step in Machine Learning (ML) and Natural Language Processing (NLP). Text classification algorithms can identify and categorize text documents based on topics, sentiment, or intent, facilitating information retrieval and NLP tasks.

There are several types of classification algorithms, each with its own strengths and weaknesses. Some of the commonly used classification algorithms include LR, RF, SVM, NB, DT, KNN and CNN.

*1) Logistic Regression (LR):* Text classification tasks are commonly performed using logistic regression. It is a statistical technique used to predict a binary outcome based on text features. Logistic regression is a versatile algorithm that can be extended in several ways to handle different types of text classification tasks. An alternative to a binary outcome can be found by extending logistic regression to include multiple classes. This can be achieved by using one logistic regression model for each class. The logistic regression model estimates the probability that a given document belongs to a particular class using a logistic (or sigmoid) function. After training, the logistic regression model can predict the probability that a new, unseen document belongs to each class. The class with the highest probability is then chosen as the predicted class for that document. It's a simple yet effective algorithm for text classification tasks, especially when the number of features (words) is large compared to the number of samples (documents).

*2) Random Forest (RF):* It is an ensemble learning algorithm that combines multiple decision trees and is known as an ensemble method. Each decision tree is a classifier that predicts the class label for a given feature vector. During training, Random Forest splits the data into subsets and selects features randomly for each split to keep the data random. Compared to individual decision trees, Random Forest results in higher accuracy and robustness by aggregating the predictions of multiple trees. By training multiple decision trees, the algorithm reduces the variance of individual trees and enhances the performance of the overall model. Random Forest is robust to overfitting and noise in the data. As a result, several subsets of data are used to train each decision tree, preventing overfitting, and improving generalizability. The algorithm automatically balances the training samples by randomly down sampling the majority class. The number of decision trees used in this analysis is 100. A small number of trees may lead to a model prone to overfitting the training data. The random state value is set to 42 in the analysis of this study. By setting a random state value, the algorithm's randomness becomes deterministic. This means it will produce the same results when run with the same data multiple times. This consistency is essential for debugging, testing, and maintaining transparency.

*3) Support Vector Machine (SVM):* It is a widely used supervised machine learning algorithm. In SVM, the kernel calculates the similarity between different data points, allowing the classifier to make predictions on new, unseen data. A SVM classifier is trained with the linear kernel, with our labelled data, where documents are categorized according to their classes. Using the labelled data, the classifier extracts patterns for classifying new documents. The training process involves solving an optimization problem called the SVM optimization problem. To minimize the rate of misidentification, a hyperplane must be found to separate the training data into different classes.

*4) Naïve Bayes (NB):* Based on Bayes' theorem, NB is a family of probabilistic classifiers. It is a simple but powerful algorithm commonly used for classification and text categorization tasks. The "naive" in Naive Bayes comes from the assumption of independence among features, meaning that the presence of a particular feature in a class is independent of the presence of other features. With word counts and frequencies as features, Multinomial Naive Bayes are often used for text classification. The training and prediction processes of Naive Bayes are computationally efficient, especially for high-dimensional datasets. For text classification tasks, Naive Bayes is often a good baseline model, despite its simplicity and naive assumption.

*5) Decision Tree (DT):* It is a supervised learning technique which organizes text data into several categories. It is a popular method in NLP and text-mining applications. A Decision Tree is a tree-like model where each node represents a decision based on the value of a particular feature. The top node is called the "root node," and the nodes that do not split further are called "leaf nodes" or "terminal nodes." Decision nodes in the tree represent conditions or questions about the input features. These conditions lead to a binary split of the data. Leaf nodes contain the output or prediction for a given subset of data. For a classification task, labels are assigned to leaf nodes according to their classes. Data splits per decision node are based on the features selected by the algorithm. There is recursion in the construction of the tree. The algorithm chooses the feature and split point at each decision node, providing the most effective separation according to the chosen criteria. When certain stopping conditions are met, such as a maximum depth or a minimum number of samples in a subset, the process continues until the completion of the subset. To make predictions for the next instance, a node is traversed from its root to its leaf node according to the input features.

*6) K-Nearest Neighbours (KNN):* An instance-based learning algorithm that uses the k-nearest neighbour algorithm makes predictions based on the feature space's k-nearest neighbours. Instead of learning a model from the training data, it memorizes the entire training dataset. The KNN algorithm uses a training set of labelled examples to create a neighbouring graph. A point in the feature space represents each example in the training set, and there is a calculation involved in determining the distance between these two points. The closest neighbours (k) to each example are then identified, and the class label of the most frequently occurring class among the neighbours is assigned to the new example. The algorithm uses a distance metric to measure the similarity between instances in the feature space. The algorithm identifies the k-nearest neighbours from the training data based on the chosen distance metric for a given test instance. The class of these neighbours is then used as a prediction for the test instance. When applying the algorithm, it is necessary to specify the number of neighbours (k). The choice of k can influence model performance. As the dataset is small, the value of k is set to 5. An odd value for k is chosen to avoid voting ties.

*7) Convolutional Neural Network (CNN):* It has been successfully applied to various text classification tasks. Text classification using CNN involves several steps, including text preprocessing, model architecture design, and training. Text data needs to be tokenized and converted into a format suitable for CNN input. They consist of multiple layers of neurons, where each layer learns different representations of the text document. The layers in a CNN typically have convolutional and pooling operations, which enable feature extraction and abstraction. Using CNNs for text classification leverages the ability of convolutional layers to capture local dependencies in text.

## 4. Experimental evaluation

In this section, the experimental measures to evaluate the dataset and the steps taken to check the results of the classification strategies is discussed. Several experiments are conducted to determine the most optimal classification model to detect child development facts and myths. The accuracy results have been investigated through the classification report for the six ML and one DL strategies is measured using several accuracy matrices, including accuracy, precision, recall, f1-score, support, macro average, weighted average, and confusion matrix, and validated using cross-validation techniques. Furthermore, the cost-effectiveness of these models has been experimented using several performance indicators. In these experiments, we utilized terms like TP, TN, FP, FN etc.

Here's a breakdown of the terms:

1. *True Positives (TP)*: The number of instances that are positive and correctly predicted as positive by the classifier.
2. *True Negative (TN)*: The number of instances that are negative and correctly predicted as negative by the classifier.
3. *False Positives (FP)*: The number of instances that are negative but incorrectly predicted as positive by the classifier.
4. *False Negatives (FN)*: The number of instances that are positive but incorrectly predicted as negative by the classifier.

### *4.1. Accuracy report*

Accuracy report provides a measure of how well the classifier or model is performing by indicating the proportion of correct predictions among the total number of cases evaluated. While accuracy is a fundamental metric, an accuracy report often includes additional metrics such as precision, recall, f1-score, support, macro average, weighted average, confusion matrix. All these matrices are used for the experiment of the models in this study.

#### *4.1.1. Accuracy*

The accuracy of a classification model is calculated as the ratio of correctly predicted instances to the total number of instances. The accuracy formula is:

$$Accuracy = \frac{Number\ of\ Correct\ Prediction}{Total\ Number\ of\ Prediction}$$

Accuracy provides an overall assessment of a model's correctness. Despite its popularity, accuracy might not be the best indicator when a class significantly outnumbers another. Precision, recall, and F1-score can provide a more nuanced evaluation of a model's performance.

*4.1.2. Precision*

Precision is one of the metrics provided in the classification report where classifier's true positive prediction ratio is equal to the total number of positive predictions it makes. Precision is calculated mathematically as follows:

$$Precision = \frac{TP}{TP + FP}$$

Precision is a measure of the accuracy of the classifier's positive predictions. It is particularly relevant when false positives are high. It indicates a low rate of false positives if the classifier has a high precision.

*4.1.3. Recall*

Recall, also known as sensitivity or true positive rate, is another metric provided in the classification report. A true positive prediction ratio is calculated by dividing the total number of actual positive instances in the dataset by the number of true positive predictions in the dataset. Recall is calculated mathematically as follows:

$$Recall = \frac{TP}{TP + FN}$$

Recall measures the classifier's ability to capture all positive instances in the dataset. It is particularly relevant when false negatives are high. A high recall indicates that the classifier has a low missing positive rate.

*4.1.4. F1 score*

The F1 score is a metric that combines precision and recall into a single value. The harmonic mean of precision and recall provides a balanced measure of classifier performance. The formula for calculating the F1 score is as follows:

$$F1 = 2 \times \frac{Precision + Recall}{Precision \times Recall}$$

In the context of the classification report provided in the code, the F1 score for each class is included in the output. It is calculated for each class based on precision and recall values. The F1 score is a valuable metric that balances precision and recall trade-offs. A higher F1 score indicates a better balance between precision and recall.

*4.1.5. Support*

Classification reports measure "support" based on the actual occurrences of each class among the specified targets. It is the number of actual instances of each class in the test set that determines the support value. Support provides context for performance metrics by providing insight into class distributions within the dataset.

*4.1.6. Macro average*

In the context of a classification report, the "macro avg" (macro average) refers to the unweighted average of precision, recall, and F1 score across all classes. It calculates the average performance across different classes without considering class imbalance.

Here's how the macro average is calculated:

1) *Precision (macro avg)*: Calculate the average precision across all classes.

$$Precision(macro\ avg) = \frac{Precision\ (Class\ 0)\ +\ Precision\ (Class\ 1)\ +\ \dots\ +\ Precision\ (Class\ N)}{N}$$

Here, precision is denoted as P and class is denoted as C.

2) *Recall (macro avg)*: Calculate the average recall across all classes.

$$Recall(macro\ avg) = \frac{Recall\ (Class\ 0)\ +\ Recall\ (Class\ 1)\ +\ ...\ +\ Recall\ (Class\ N}{N}$$

3) *F1-score (macro avg)*: Calculate the average F1-score across all classes.

$$F1(macro\ avg) = \frac{F1 - score\ (Class\ 0)\ +\ F1 - score\ (Class\ 1)\ +\ ...\ +\ F1 - score\ (Class\ N)}{N}$$

Here, N is the total number of classes, P is precision, R is recall, and C is class.

The macro average treats all classes equally, regardless of their size in the dataset. It is useful when you want to evaluate the overall performance without being biased by imbalances in class distribution.

*4.1.7. Weighted average*

In the context of a classification report, the "weighted average" refers to an average of precision, recall, and F1-score across different classes, where each class contribution is weighted by its support (the number of true instances for each class). This type of averaging is particularly useful when there is class imbalance, as it considers the contribution of each class proportionate to its prevalence in the dataset.

Here's how the weighted average is calculated:

1) *Weighted Precision*: Calculate the weighted average precision across all classes.

$$Precision\ (weighted\ avg) = \frac{P\ (Class\ 0)\ \times\ S\ (Class\ 0)\ +\ P\ (Class\ 1)\ \times\ S\ (Class\ 1)\ +\ ...\ +\ P\ (Class\ N)\ \times\ S\ (Class\ N)}{Total\ Support}$$

2) *Weighted Recall*: Calculate the weighted average recall across all classes.

$$Recall\ (weighted\ avg) = \frac{Recall\ (Class\ 0) \times\ S\ (Class\ 0) +\ Recall\ (Class\ 1) \times\ S\ (Class\ 1) +\ ...\ +\ Recall\ (Class\ N) \times\ S\ (Class\ N)}{Total\ Support}$$

3) *Weighted F1-score*: Calculate the weighted average F1-score across all classes.

$$F1(weighted\ avg) = \frac{F(Class\ 0)\ \times S\ (Class\ 0)\ +\ F\ (Class\ 1)\ \times\ S\ (Class\ 1)\ +\ ...\ +\ F(Class\ N)\ \times\ S\ (Class\ N)}{Total\ Support}$$

Here, Total Support is the sum of support values for all classes, P is precision, F is F1-score, and S is support.

The weighted average gives more importance to the performance of classes with larger support, providing a balanced evaluation in the presence of imbalanced class distributions.

*4.1.8. Confusion matrix*

A confusion matrix is a table used in classification to assess the performance of a classification algorithm. It summarizes the counts of the classifier's TP, TN, FP, and FN predictions. Each row of the matrix represents instances in an actual class, while each column represents instances in a predicted class. The confusion matrix provides a detailed breakdown of the model's performance, particularly regarding the types of errors it makes.

*4.2. Cross-fold validation*

Cross-validation is a resampling technique used in machine learning to assess predictive model performance. The main goal of cross-validation is to use the available data in an efficient way, especially when the dataset is limited. It helps in estimating how the performance of a model will generalize to an independent dataset. Cross-validation is useful for assessing a model's generalization performance, identifying potential issues like overfitting or underfitting and tuning hyperparameters. The stratified k-fold cross-validation in machine learning is used to deal with imbalanced datasets.

*4.2.1. k-fold cross validation*

k-fold cross-validation is the most popular form of cross-validation. The dataset is randomly partitioned into k equally sized folds using this technique. The model is trained *k* times, each using *k*−1 folds for training and the remaining fold for validation. This process is repeated *k* times, and the performance measures (such as accuracy, precision, recall, etc.) are averaged over the *k* folds.

Steps of *k*-fold cross-validation:

1) *Divide the dataset*: Randomly split the dataset into *k* equally sized folds.
2) *Model Training and Validation*: For *i*=1 to *k*:

Use the $i^{th}$ fold as the validation set.

- Train the model on the remaining *k*−1 folds.
- Evaluate the model on the validation set.

3) *Performance Measure*: Calculate the average performance measure (e.g., accuracy) over the *k* iterations.

This process ensures that each data point is used for validation exactly once. In addition, the model's performance is estimated over a wide range of validation sets, which produces a more robust estimate.

### *4.2.2. Leave-one-out validation*

LOO (Leave-One-Out) cross-validation is a particular case of *k*-fold cross-validation where *k* equals the number of samples in the dataset. The dataset is partitioned into n folds under LOO cross-validation, where *n* is the number of samples. One sample is used as a validation set for each iteration, and the model is trained on the remaining *n*−1 samples. This process is repeated *n* times, with each sample being used once as a validation set.

Steps of LOO cross-validation: *Divide the dataset*: For each sample *i*=1 to *n*:

- Use sample *i* as the validation set.
- Training the model on the remaining *n*−1 samples.
- Evaluate the model on the validation set containing only samples *i*.

2) *Performance Measure*: Calculate the average performance measure (e.g., accuracy) over the *n* iterations.

LOO cross-validation thoroughly assesses a model's performance because it uses only one sample for training in each iteration. It is, however, computationally intensive, especially when it comes to large datasets, because n models need to be trained. Despite the computational cost, LOO cross-validation is helpful in situations where maximizing the use of available data for training is crucial, and computational resources allow for repeated model training on almost the entire dataset.

### *4.3. Cost-effectiveness evaluation*

Evaluation of cost-effectiveness of the models for different feature extraction techniques is done by using the performance indicator: testing time per statement. Testing time per statement is the time taken for a model to make predictions on a single instance of data. The equation to calculate testing time per statement is,

$$\text{Testing time per statement} = \text{Testing Time} / \text{Number of statements}$$

Where, testing time is the time taken to perform the predictions (testing_time) and number of Statements is the total number of statements (or predictions) in the test set.

## 5. Evaluation results

In this section, the data analysis tools, and the different visualizations of the experiment results is discussed. The purpose of this paper is to examine myths and facts about child development using text mining techniques and classification models, providing valuable insights derived from newly gathered data from publicly available websites. In this study, empirical data was categorized into four functional labels, and different aspects were examined.

### *5.1. Data analysis tools*

For the experiment, Python is used to write codes and Jupiter Notebook is used to execute them with the help of several Python libraries. Several packages is used in Python like nltk, scikit, numpy, pandas, word_tokenize, stopwords, PorterStemmer, WordNetLemmatizer, pos_tag, CountVectorizer, TfidfVectorizer, MultinomialNB, SVC, kNeighborsClassifier, DecisionTreeClassifier, LinearRegression, LogisticRegression, RandomForestClassifier, mean_squared_error, accuracy_score, classification_report, confusion_matrix, cross_val_score, LeaveOneOut, matplotlib.pyplot.

Natural Language Toolkit (NLTK) is an open-source library that contains a range of tools for working with natural language processing (NLP) tasks. It provides classes and functions for tasks such as tokenization, stemming, tagging, parsing, and sentiment analysis. scikit-learn is an open-source library for machine learning in Python. It is built on top of NumPy and SciPy and provides a range of algorithms for supervised and unsupervised learning tasks. In addition to classification, regression, clustering, and dimensionality reduction, scikit-learn is widely used for other tasks as well.

NumPy is a numerical computing library for Python. It provides an array-based data structure, along with functions to perform mathematical operations on arrays. NumPy is widely used for scientific computing, data analysis, and machine learning tasks. Pandas is a data structures library for Python, often used for data analysis and manipulation. It provides a range of data structures, such as series, data frames, and panel objects, along with powerful tools for data cleaning, reshaping, and data visualization.

The word_tokenizefunction in NLTK splits a sentence into its individual tokens, also known as words. It treats punctuation marks and special characters as individual tokens. The stopwords function in NLTK removes common words such as articles, prepositions, and pronouns from a sentence. This helps in reducing noise and making the text more representative and meaningful. The PorterStemmer class in NLTK is a stemming algorithm that reduces the inflected form of a word to its root form. This helps in simplifying the vocabulary and improving the accuracy of text analysis tasks.

The pos_tag function in NLTK annotates a sentence with its part-of-speech tags. This helps us understand the grammatical structure of the sentence and identify different word classes. The WordNetLemmatizer class in NLTK lemmatizes words by retrieving their lemmas from a WordNet database. This helps disambiguate words with similar forms and improves the text's understanding.

Matplotlib is an open-source plotting library for Python widely used to create informative visualizations. It provides various plotting and charting functionalities, including line plots, scatter plots, and histograms. The CountVectorizer class in scikit-learn is a transformer that converts text data into a bag-of-words representation. It counts the frequency of each unique word and stores them in a high-dimensional vector. The TfidfVectorizer class in scikit-learn is a transformer that converts text data into a TF-IDF representation. It weights each word by its frequency and its inverse document frequency.

The MultinomialNB class in scikit-learn is a probabilistic classifier for multi-class classification tasks. It assumes that the classes are mutually exclusive and exhaustive. The SVC class in scikit-learn is a support vector classifier. It classifies instances based on the maximum margin hyperplane that separates the classes. The kNeighborsClassifier class in scikit-learn is a non-parametric classifier that assigns instances to classes based on the majority vote of their neighbours. The DecisionTreeClassifier class in scikit-learn is a supervised learning algorithm that builds a binary classification tree from the training data. It recursively partitions the data into smaller and smaller sets, until each set contains only one class. The LogisticRegression class in scikit-learn is a machine learning algorithm for binary classification tasks. It maps input features to a binary classification decision based on the sigmoid function. The RandomForestClassifier class in scikit-learn is an ensemble learning algorithm that combines multiple decision trees to improve classification accuracy. It trains each decision tree on a subset of the training data and combines their predictions to obtain the final classification. The LinearRegression class in scikit-learn is a supervised learning algorithm for linear regression problems. Predicted and actual values are compared to minimize the squared errors.

The mean squared error (MSE) is a metric used to evaluate the performance of a regression model. It quantifies the average squared difference between the predicted and actual values. The accuracy score is a metric used to evaluate the performance of a classification model. It quantifies the proportion of correctly classified instances. The classification report is a visualization tool that overviews a classification model's performance. It provides precision, recall, F1-score, and confusion matrix metrics. The confusion matrix is a matrix that represents the predicted and actual classes for each observation. It helps in understanding the model's performance and the misclassification errors. When a classification model is trained on unseen data, it is evaluated via a cross-validation score. It measures the average classification accuracy over multiple rounds of cross-validation. The Leave-One-Out cross-validation technique involves evaluating the performance of a classification model on each observation in the training data, one at a time, and averaging the results.

### *5.2. Results*

The data was pre-processed using various text mining techniques, including tokenization, stop word removal, and stemming. These techniques help to extract meaningful information from unstructured text data and prepare it for analysis. After preprocessing, the word cloud was generated from the text data. Fig. 2 represents the word cloud and reveals several noteworthy findings. One widely used technique for analysing word distributions is the word cloud, which allows us to visually represent the frequency of each word in a dataset. The most frequently occurring word in the data is 'develop'. It is positioned centrally and highlighted in a bold font, emphasizing its significance in the data. 'learn', 'brain', and 'play' have been found following close behind, which are depicted as significant words, positioned near 'develop'.

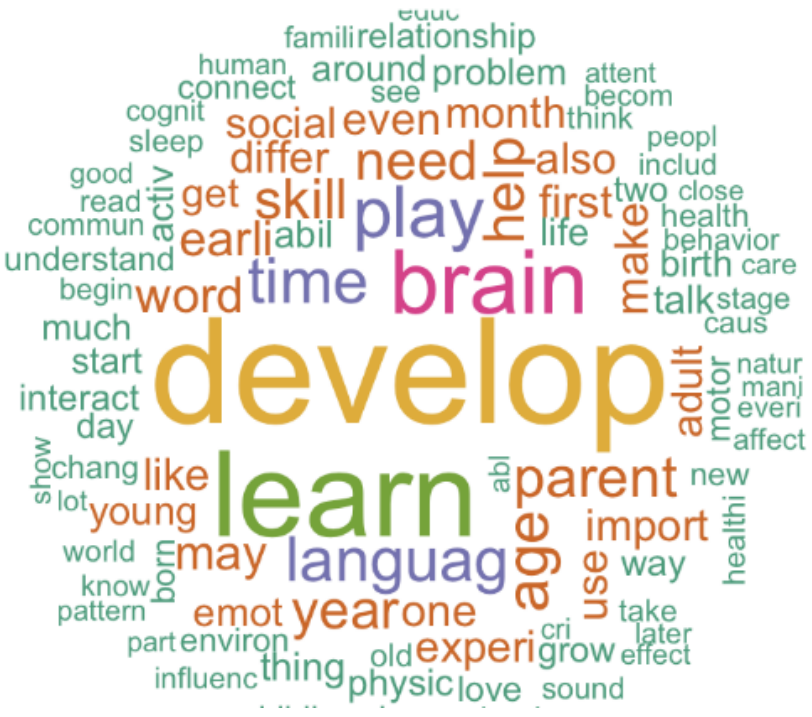

**Figure 2.** Word cloud of preprocessed data.

The structured data was analysed using six traditional classification models: LR, RF, SVM, NB, DT, KNN, and CNN. The experiment investigated the effect of two feature extraction techniques on the accuracy rates of the AI models. The classification report is described in Table 2 while 80% of the data was used for training, and 20% of the data was used for testing from the whole dataset. The classification models have been tested for two labels: Fact and Myth.

The experiment results show that LR has the highest accuracy rate of 90% with BoW, while CNN has the highest accuracy rate of 89% with TF-IDF. On the other hand, NB has the lowest accuracy rate (74%) in TF-IDF, and KNN has the lowest accuracy rate (70%) in BoW. For Fact, the highest precisions were achieved by LR (95%) using BoW, CNN (92%) using TF-IDF for facts, and the lowest precision was achieved by NB (73% for TF-IDF and 85% for BoW). For Myth, NB has the highest precision (90% with BoW and 94% with TF-IDF), and the lowest precision held by KNN (51%) for BoW, and DT (64%) for TF-IDF. For facts, NB had a 99% recall through TF-IDF and 97% by BoW, which are the best models among other models. LR and NB have the highest macro average precision (87%) with BoW. LR had the optimum F1-score (90%) based on BoW feature extraction for facts, myths, macro average and weighted average. In contrast, CNN obtained these best values (89%) using TF-IDF. Support is 192 for Facts and 87 for Myths.

**Table 2.** Classification report for 80% training and 20% testing.

| Classifications | Labels | TF-IDF | | | | | BoW | | | | |
|---|---|---|---|---|---|---|---|---|---|---|---|
| | | Precision | Recall | F1-score | Support | Accuracy | Precision | Recall | F1-score | Support | Accuracy |
| LR | Fact | 0.83 | 0.97 | 0.90 | 192 | 0.85 | **0.95** | 0.90 | **0.92** | 192 | **0.90** |
| | Myth | 0.89 | 0.57 | 0.70 | 87 | | 0.80 | **0.89** | **0.84** | 87 | |
| | Macro avg | 0.86 | 0.77 | 0.80 | 279 | | **0.87** | **0.89** | **0.88** | 279 | |
| | Weighted avg | 0.85 | 0.85 | 0.83 | 279 | | **0.90** | **0.90** | **0.90** | 279 | |
| RF | Fact | 0.88 | 0.90 | 0.89 | 192 | 0.85 | 0.88 | 0.90 | 0.89 | 192 | 0.85 |
| | Myth | 0.76 | 0.74 | 0.75 | 87 | | 0.77 | 0.72 | 0.75 | 87 | |
| | Macro avg | 0.82 | 0.82 | 0.82 | 279 | | 0.82 | 0.81 | 0.82 | 279 | |
| | Weighted avg | 0.84 | 0.85 | 0.85 | 279 | | 0.84 | 0.85 | 0.84 | 279 | |
| SVM | Fact | 0.87 | 0.96 | 0.91 | 192 | 0.87 | 0.93 | 0.85 | 0.89 | 192 | 0.85 |
| | Myth | 0.88 | 0.69 | 0.77 | 87 | | 0.72 | 0.86 | 0.79 | 87 | |
| | Macro avg | **0.88** | 0.82 | 0.84 | 279 | | 0.83 | 0.86 | 0.84 | 279 | |
| | Weighted avg | 0.88 | 0.87 | 0.87 | 279 | | 0.87 | 0.85 | 0.86 | 279 | |
| NB | Fact | 0.73 | **0.99** | 0.84 | 192 | 0.74 | 0.85 | **0.97** | 0.91 | 192 | 0.86 |
| | Myth | **0.94** | 0.18 | 0.31 | 87 | | **0.90** | 0.62 | 0.73 | 87 | |
| | Macro avg | 0.84 | 0.59 | 0.57 | 279 | | **0.87** | 0.79 | 0.82 | 279 | |
| | Weighted avg | 0.80 | 0.74 | 0.67 | 279 | | 0.87 | 0.86 | 0.85 | 279 | |
| DT | Fact | 0.87 | 0.82 | 0.84 | 192 | 0.79 | 0.88 | 0.82 | 0.85 | 192 | 0.80 |
| | Myth | 0.64 | 0.72 | 0.68 | 87 | | 0.66 | 0.76 | 0.71 | 87 | |
| | Macro avg | 0.76 | 0.77 | 0.76 | 279 | | 0.77 | 0.79 | 0.78 | 279 | |
| | Weighted avg | 0.80 | 0.79 | 0.79 | 279 | | 0.81 | 0.80 | 0.81 | 279 | |
| KNN | Fact | 0.90 | 0.91 | 0.90 | 192 | 0.87 | 0.91 | 0.62 | 0.74 | 192 | 0.70 |
| | Myth | 0.79 | 0.78 | 0.79 | 87 | | 0.51 | 0.86 | 0.64 | 87 | |
| | Macro avg | 0.85 | 0.84 | 0.85 | 279 | | 0.71 | 0.74 | 0.69 | 279 | |
| | Weighted avg | 0.87 | 0.87 | 0.87 | 279 | | 0.78 | 0.70 | 0.71 | 279 | |
| CNN | Fact | **0.92** | 0.92 | **0.92** | 192 | **0.89** | 0.93 | 0.89 | 0.91 | 192 | 0.88 |
| | Myth | 0.82 | **0.83** | **0.82** | 87 | | 0.78 | 0.86 | 0.82 | 87 | |
| | Macro avg | 0.87 | **0.87** | **0.87** | 279 | | 0.86 | 0.88 | 0.87 | 279 | |
| | Weighted avg | **0.89** | **0.89** | **0.89** | 279 | | 0.89 | 0.88 | 0.88 | 279 | |

The accuracy result is observed with the confusion matrix, as shown in Fig. 3 and 4. Fig. 3 describes the confusion matrix for classification accuracy using TF-IDF, while Fig. 4 explains the confusion matrix for classification accuracy utilising Bow.

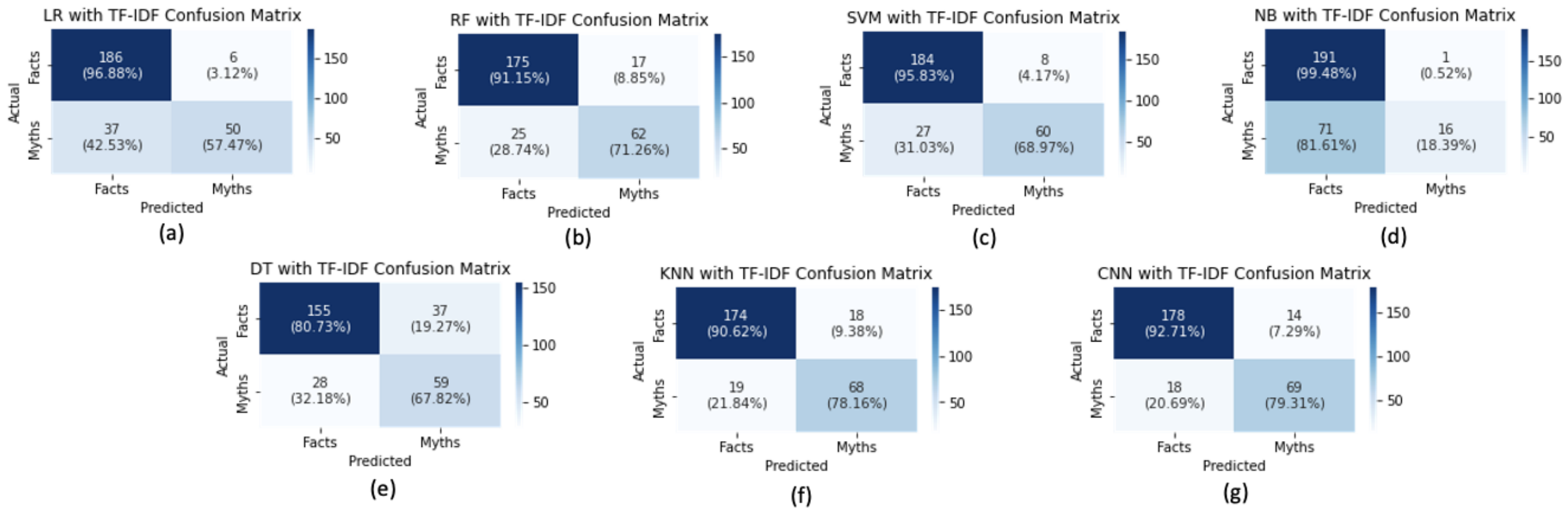

**Figure 3.** Confusion matrix for TF-IDF for 80% training and 20% testing.

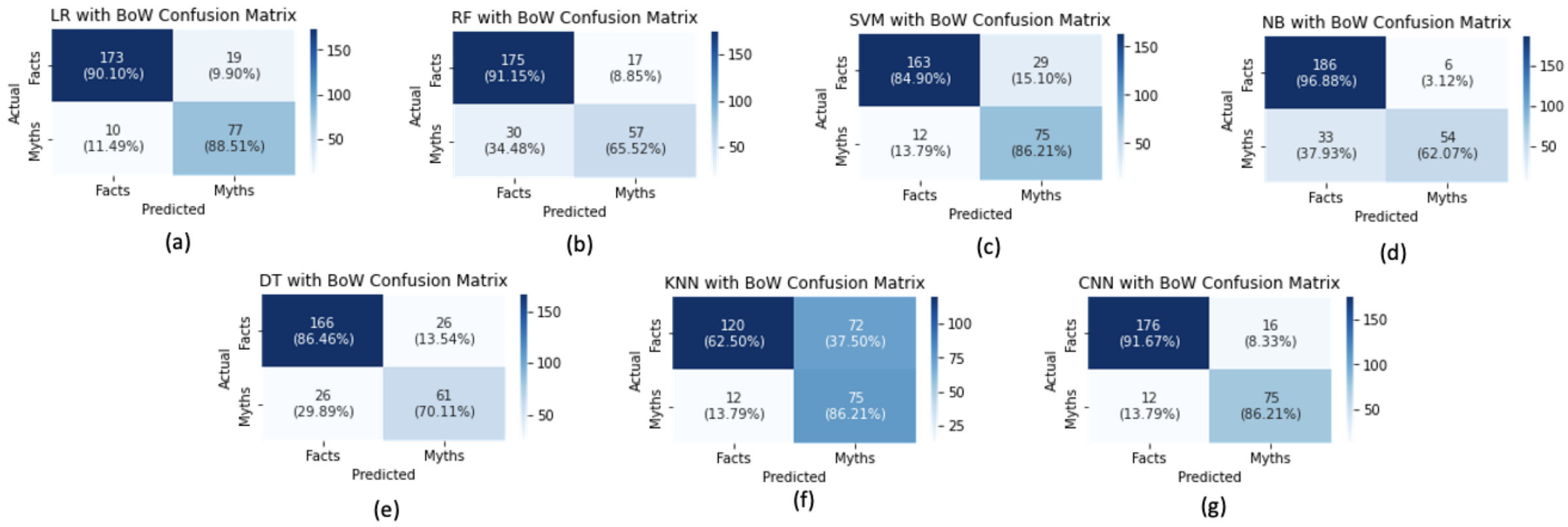

**Figure 4.** Confusion matrix for BoW for 80% training and 20% testing.

Table 3 reports the accuracy results of all seven AI methods across different training-testing splits (80-20, 70-30, 60-40). LR shows strong performance, particularly with BoW, achieving its highest accuracy of 0.90 with an 80-20 split. CNN model consistently performs well across all splits and feature extraction methods, maintaining high accuracy scores (ranging from 0.88 to 0.90). Conversely, the Naive Bayes model exhibits a significant drop in accuracy with TF-IDF compared to BoW, indicating it might not handle the TF-IDF features as effectively. KNN demonstrates notable variability, particularly under BoW, where its accuracy ranges from 0.67 to 0.73, but it performs exceptionally well with TF-IDF, achieving an accuracy of up to 0.87. The Decision Tree model, however, maintains relatively lower and more consistent accuracy across all scenarios.

**Table 3.** Accuracy for different training and testing set.

| **Training Testing percentage** | **80-20** | | **70-30** | | **60-40** | |
|---|---|---|---|---|---|---|
| **Models** | BoW | TF-IDF | BoW | TF-IDF | BoW | TF-IDF |
| **LR** | **0.90** | 0.85 | **0.89** | 0.84 | 0.86 | 0.82 |
| **RF** | 0.85 | 0.85 | 0.85 | 0.84 | 0.83 | 0.84 |
| **SVM** | 0.85 | 0.87 | 0.86 | 0.85 | 0.84 | 0.83 |
| **NB** | 0.86 | 0.74 | 0.84 | 0.73 | 0.82 | 0.72 |
| **DT** | 0.80 | 0.79 | 0.78 | 0.74 | 0.78 | 0.79 |
| **KNN** | 0.70 | 0.87 | 0.73 | 0.86 | 0.67 | 0. 83 |
| **CNN** | 0.88 | **0.89** | **0.89** | **0.89** | **0.89** | **0.90** |

Different Cross-validation techniques precisely 5-fold, 10-fold and LOO have been used to analyse the accuracy of the models with the data. As depicted in Table 4, the experiment result indicates that LR shows a marked improvement when using BoW, reaching up to 86% accuracy with Leave One Out validation, compared to 82% with TF-IDF. An interesting feature is the consistency in the performance of the CNN model, which achieves high accuracy scores (86% to 87%) regardless of the cross-validation technique used. Naive Bayes also performs significantly better with BoW, achieving up to 84% accuracy, whereas with TF-IDF, its performance is noticeably lower (around 72% to 73%). The KNN model, in contrast, performs better with TF-IDF, achieving consistent scores around 82% to 83%, while it shows poorer performance with BoW, particularly with 5-Fold and 10-Fold validation. The Decision Tree model maintains relatively stable and lower accuracy scores across both feature extraction methods and validation techniques. The SVM and RF models exhibit comparable performance across both feature extraction methods, with slight variations depending on the validation technique.

**Table 4.** Cross-fold validation for 80% Training-set and 20% Testing-set.

| Feature Extraction | Classifications | Cross Fold Validation | | |
|---|---|---|---|---|
| | | K fold | | Leave one out |
| | | 5-Fold | 10-Fold | |
| BoW | **LR** | 0.84 | **0.85** | **0.86** |
| | **RF** | 0.83 | 0.83 | 0.83 |
| | **SVM** | 0.83 | 0.83 | 0.83 |
| | **NB** | 0.82 | 0.83 | 0.84 |
| | **DT** | 0.74 | 0.75 | 0.75 |
| | **KNN** | 0.67 | 0.67 | 0.68 |
| | **CNN** | **0.85** | **0.85** | 0.85 |
| TF-IDF | **LR** | 0.81 | 0.81 | 0.82 |
| | **RF** | 0.83 | 0.83 | 0.82 |
| | **SVM** | 0.83 | 0.84 | 0.84 |
| | **NB** | 0.72 | 0.72 | 0.73 |
| | **DT** | 0.75 | 0.75 | 0.76 |
| | **KNN** | 0.83 | 0.83 | 0.82 |
| | **CNN** | **0.87** | **0.86** | **0.85** |

As depicted in Table 5, a performance indicator has been experimented with to analyse the models' performance for the data. This table highlights the trade-offs between computational efficiency and model complexity, emphasizing the cost-effectiveness of LR and NB, and the high computational demands of CNN models. For both feature extraction methods, LR excels in speed and efficiency by maintaining a moderate testing time per statement (0.97 µs). NB also performs efficiently, particularly with TF-IDF, achieving rapid testing time per statement (0.89 µs). Conversely, KNN shows a significant increase in testing time per statement with TF-IDF (3205.30 µs), indicating inefficiency with this feature extraction method. The CNN model, while effective in terms of accuracy, demonstrates much higher computational costs, resulting in the highest testing time per statement (1916.16 µs).

**Table 5.** Evaluation of cost-effectiveness for 80% training and 20% testing.

| Methods | Models | Feature Techniques | Testing time per Statement (µs) |
|---|---|---|---|
| ML | LR | BoW | **0.97** |
| | | TF-IDF | **0.97** |
| | RF | BoW | 34.26 |
| | | TF-IDF | 36.70 |
| | SVM | BoW | 92.44 |
| | | TF-IDF | 142.54 |
| | NB | BoW | 1.54 |
| | | TF-IDF | **0.89** |
| | DT | BoW | 1.61 |
| | | TF-IDF | 2.47 |
| | KNN | BoW | 113.08 |
| | | TF-IDF | 3205.30 |
| DL | CNN | BoW | 1502.65 |
| | | TF-IDF | 1916.16 |

## 6. Discussion

With the advent of the digital age, there has been an increase in misinformation. False information can easily be disseminated on social media and online platforms, causing harm to individuals and communities. Various techniques have been developed to detect and identify misinformation to combat this issue. Child development myths and facts have not been researched except for one study on child abuse cases [7]. So, any relevant comparison with a previous work is not possible. Text-mining techniques is used to analyse child development facts and myths and to check the accuracy of six traditional classification models and one deep learning model using two feature extraction methods. The methodological steps have been demonstrated in Fig. 1.

Some examples of data have been provided in Table 1. The data has been cleaned, tokenized, stemmed, and lemmatized as part of the preprocessing, valuable insights were gained into the data, and these insights are illustrated by a word cloud in Fig. 2. Word clouds provide valuable insights into the content of a piece of information using visual analysis. Identifying patterns and themes in our data is easier when we create a visual representation of the most frequently used words. The word cloud shows that 'develop' is the most used word in the data by emphasizing significance while placing it in the center and bolding its font. Similar analysis results have been obtained in [59, 64] using word cloud. Additionally, it is observed that 'learn', 'play', and 'brain' all display significant words near 'develop'. These are interesting findings, and these words are somehow interconnected [65].

The data is analysed using six robust ML classifiers and one deep learning model using two feature extraction techniques for three different training-testing sets. The experiment results are shown in Table 2 and 3, where Table 2 describes the accuracy report for 80% training set and 20% testing set. Table 3 depicts the accuracy result for three different training-testing sizes. Moreover, the cross-validation methods are utilised for evaluating the accuracy. Table 4 illustrates the accuracy results of the models using 5-fold, 10-fold, and LOO for both feature extraction techniques. Furthermore, the cost-effectiveness of the models has been investigated in Table 5.

According to the result, LR demonstrated the greatest accuracy, achieving 90% with BoW feature extraction. It is significantly higher than the 70% accuracy achieved by KNN. This finding of LR accuracy is consistent with other studies [66, 67], although they used Twitter data for fake news detection. As a result of its exceptional speed and efficiency, LR processes with low testing times per statement (0.97 μs) for both feature extraction methods. The results of this study suggest that LR combined with BoW is particularly effective at accurately categorizing child development information, hence providing a valuable tool for combating misinformation. This recommendation is somehow supported by a study on message classification [66].

While this study provides valuable insights into child development and misinformation analysis, there are a few limitations to be noted. The dataset may not comprehensively represent all sources, and the employed feature extraction techniques and models could have inherent limitations. Besides, applying these models in real-world contexts requires additional enhancement, user testing, and integration (e.g., with relevant websites or apps) efforts.

## 7. Conclusion

Misinformation has increased with the advent of the digital age. Online platforms and social media can easily disseminate false information, which harms communities and individuals. Parents often use the Internet to research issues they don't understand, including myths, in today's data-centric era. Parents may be confused by the myths, negatively affecting how their children are treated after learning them. The fight against misinformation has been aided by the development of numerous techniques. Except for one study on child abuse cases, no studies have been conducted on child development myths and facts. This paper presents a novel approach to analyse myths and facts related to child development using text-mining techniques and classification models.

By harnessing the power of NLP and ML, meaningful information is extracted from newly gathered data, enabling us to gain valuable insights into the topic. The empirical data in this study has been categorized into two functional labels, and different aspects have been explored. Several pre-processing methods is done to get the effective analysis results from the data. In the next step, the structured data was analysed using two feature extraction method and six traditional classification models and one deep learning model, with LR achieving the highest accuracy rate using BoW. Cross-validation was performed using k-fold and loo methods, providing confidence in our findings. Cost-effectiveness evaluation is done using a performance indicator. Among them, LR achieves the highest accuracy rate using BoW with overall optimum testing time and robust testing time per statement. In addition to shedding light on myths and facts regarding child development, this study provides a solid base for future research in the field.

## CRediT authorship contribution statement

**Mehedi Tajrian:** Conceptualization, Data curation, Formal analysis, Visualization, Validation, Software, Resources, Methodology, Investigation, Writing – original draft. **Azizur Rahman:** Conceptualization, Writing – review & editing, Supervision. **Muhammad Ashad Kabir:** Writing – review & editing, Supervision, Conceptualization. **Rafiqul Islam:** Writing – review & editing, Supervision.

## Declaration of competing interest

The authors declare no competing interests.

## Data availability statement

The data used in this study are openly available at https://researchoutput.csu.edu.au/en/datasets/myths-and-facts-data-on-child-development.